\documentclass[runningheads]{llncs}
\usepackage{graphicx}
\usepackage{comment}
\usepackage{amsmath,amssymb} 
\usepackage{color}

\usepackage{epsfig}
\usepackage{bm}
\usepackage{multirow}
\usepackage{booktabs}
\usepackage{graphicx}
\usepackage{wrapfig}
\usepackage{enumerate}

\begin{document}
\pagestyle{headings}
\mainmatter
\def\ECCVSubNumber{2287}  

\title{Graph-Based Social Relation Reasoning} 

\titlerunning{Graph-Based Social Relation Reasoning}
\author{Wanhua Li\inst{1,2,3} \and
Yueqi Duan\inst{1,2,3,5,}$^{\dagger}$ \and
Jiwen Lu\inst{1,2,3,}$^{\ast}$ \and
Jianjiang Feng\inst{1,2,3} \and
Jie Zhou\inst{1,2,3,4}}
\authorrunning{W. Li et al.}
\institute{Department of Automation, Tsinghua University, China
\and State Key Lab of Intelligent Technologies and Systems, China
\and Beijing National Research Center for Information Science and Technology, China
\and Tsinghua Shenzhen International Graduate School, Tsinghua University, China
\and Stanford University \\
\email{li-wh17@mails.tsinghua.edu.cn,duanyq19@stanford.edu,
\{lujiwen,jfeng,jzhou\}@tsinghua.edu.cn}
}

\maketitle

\let\thefootnote\relax\footnotetext{$^{\dagger}$ Work done while at Tsinghua University.}
\let\thefootnote\relax\footnotetext{$^{\ast}$ Corresponding author}

\begin{abstract}
Human beings are fundamentally sociable --- that we generally organize our social lives in terms of relations with other people. Understanding social relations from an image has great potential for intelligent systems such as social chatbots and personal assistants. In this paper, we propose a simpler, faster, and more accurate method named graph relational reasoning network (GR$^2$N) for social relation recognition. Different from existing methods which process all social relations on an image independently, our method considers the paradigm of jointly inferring the relations by constructing a social relation graph. Furthermore, the proposed GR$^2$N constructs several virtual relation graphs to explicitly grasp the strong logical constraints among different types of social relations. Experimental results illustrate that our method generates a reasonable and consistent social relation graph and improves the performance in both accuracy and efficiency.
\end{abstract}


\section{Introduction}

Social relations are the theme and basis of human life, where most human behaviors occur in the context of the relationships between individuals and others \cite{reis2000relationship}.
The social relation is derived from human social behaviors and defined as the association between individual persons. Social relation recognition from images is in the ascendant in the computer vision community \cite{sun2017domain,li2017dual}, while social relationships have been studied in social psychology for decades \cite{conte1981circumplex,fiske1992four,bugental2000acquisition}. There has been a growing interest in understanding relations among persons in a given still image due to the broad applications such as potential privacy risk warning \cite{sun2017domain}, intelligent and autonomous systems \cite{Wang2018Deep}, and group activity analysis \cite{lan2012social}.

\begin{figure}[t]
  \centering
  \includegraphics[width=1.0\linewidth]{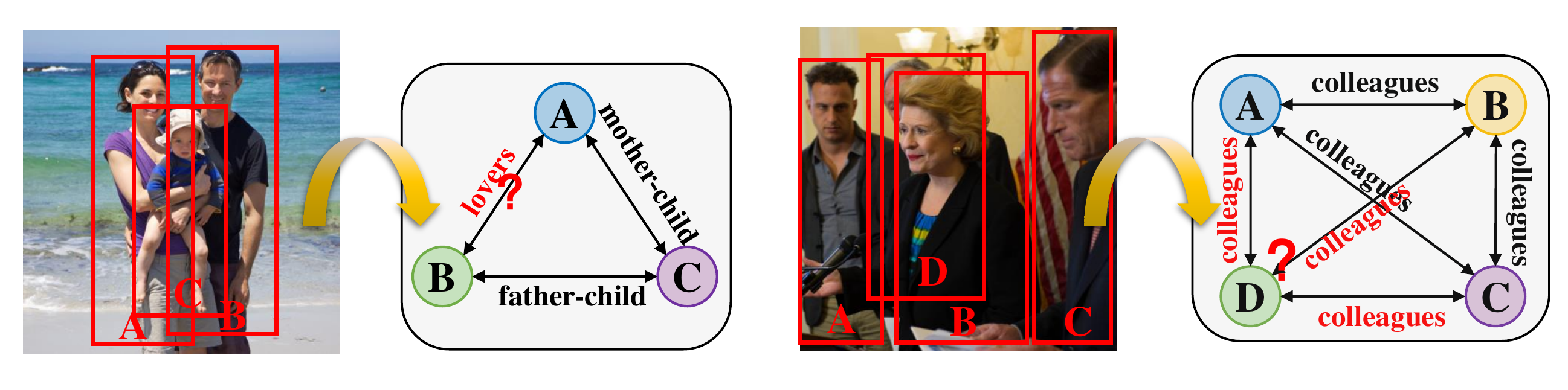}
  \caption{ Examples of how the relations on the same image help each other in reasoning. We observe that social relations on an image usually follow strong \emph{logical constraints}.}
  \label{fig:introduction}
\end{figure}

In a social scene, there are usually many people appearing at the same time, which contains various social relations.
Most existing methods recognize social relations between person pairs separately \cite{sun2017domain,li2017dual,Wang2018Deep,Arushi2019social},
where each forward pass only processes a pair of bounding boxes on an image.
However, as social relations usually form a reasonable social scene, they  are not independent of each other, but highly correlated instead.
Independently predicting the relations on the same image suffers from the high locality in social scenes,
which may result in an unreasonable and contradictory social relation graph.
To this end, we consider that jointly inferring all relations for each image helps  construct a reasonable and consistent social relation graph with a thorough understanding of the social scene.
Moreover, as social relations on the same image usually follow strong \emph{logical constraints}, simultaneously considering all relations can effectively exploit the consistency of them.
Two examples are shown in Fig. \ref{fig:introduction}. In the first example, when we know that the relation between A and C is mother-child and that between B and C is father-child, we easily infer that A and B are lovers.
In the second example, we can quickly understand the social scene through the relations among A, B, and C, and infer the relations between D and others, even if D is heavily occluded. Clearly, the relations on the same image help each other in reasoning, which is not exploited in existing methods as an important cue.


In this paper, we propose a graph relational reasoning network (GR$^2$N) to jointly infer all relations on an image with the paradigm of treating images as independent samples rather than person-pairs. In fact, the image-based paradigm is closer to the nature of human perception of social relations since human beings do not perceive the social relations of a pair of people in isolation. Many variants of graph neural networks (GNNs) can collaborate with the image-based paradigm by building a graph for each image, where the nodes represent the persons and the edges represent the relations. However, for social relation recognition, different kinds of social relations have strong logical constraints as shown in Fig. \ref{fig:introduction}. Most existing GNNs' methods simply exploit contextual information via message passing, which fails to explicitly grasp the logical constraints among different types of social relations.
To exploit the strong logical constraints, the proposed GR$^2$N constructs different virtual relation graphs for different relation types with shared node representations. Our method learns type-specific messages on each virtual relation graph and updates the node representations by aggregating all neighbor messages across all virtual relation graphs. In the end, the final representations of nodes are utilized to predict the relations of all pairs of nodes on the graph.

To summarize, the contributions of this paper are three-fold:
\begin{itemize}
\item
  To the best of our knowledge, our method is the first attempt to jointly reason all relations on the same image for social relation recognition. Experimental results verify the superiority of the paradigm shift from person pair-based to image-based.

\item
  We proposed a \emph{simper} method named GR$^2$N to collaborate with the image-based paradigm, which constructs several virtual relation graphs to explicitly model the logical constraints among different types of social relations. GR$^2$N only uses \emph{one input image patch} and \emph{one convolutional neural net} while the existing approaches usually use many more patches and nets.

\item
  The proposed GR$^2$N is \emph{faster} and \emph{more accurate.} Unlike existing methods that only handle the relation of one person pair at a time, our method processes all relations on an image at the same time with a single forward pass, which makes our method computationally efficient. Finally, our method is $2 \times  \sim  7 \times$ faster than other methods due to the paradigm shift. The proposed GR$^2$N effectively exploits the information of relations on the same image to generate a reasonable and consistent social relation graph. Extensive experimental results on the PIPA and PISC datasets demonstrate that our approach outperforms the state-of-the-art methods.
\end{itemize}

\section{Related Work}
\textbf{Social Relationship Recognition.}
Human face and body images contain rich information~\cite{lu2013cost,lu2014reconstruction,wang2014person,deng2019arcface,li2019bridgenet}.
In recent years, social relationship recognition from an image has attracted increasing interest in the computer vision community \cite{sun2017domain,li2017dual,Wang2018Deep,Arushi2019social}. Some early efforts to mine social information include kinship recognition \cite{li2020graph,lu2013neighborhood,wang2010seeing,lu2017discriminative}, social role recognition \cite{ramanathan2013social,shu2015joint} and occupation recognition \cite{shao2013you,song2011predicting}. Lu \emph{et al.} \cite{lu2013neighborhood} proposed a neighborhood repulsed metric learning (NRML) method for kinship verification. Wang \emph{et al.} \cite{wang2010seeing} explored using familial social relationships as a context for recognizing people and  the social relationships between pairs of people. Ramanathan \emph{et al.} \cite{ramanathan2013social} presented  a conditional random field method to recognize social roles played by people in an event. Personal attributes and contextual information were usually used in occupation recognition \cite{shao2013you}. Zhang \emph{et al.} \cite{zhang2018facial} devised an effective multitask network to learn social relation traits from face images in the wild. Sun \emph{et al.} \cite{sun2017domain} extended the PIPA database by 26,915 social relation annotations based on the social domain theory \cite{bugental2000acquisition}. Li \emph{et al.} \cite{li2017dual} collected a new social relation database and proposed a dual-glance model, which explored useful and complementary visual cues. Wang \emph{et al.} \cite{Wang2018Deep} proposed a graph reasoning model to incorporate common sense knowledge of the correlation between social relationships  and semantic contextual cues. Arushi \emph{et al.} \cite{Arushi2019social} used memory cells like GRUs to combine human attributes and contextual cues. However, these methods deal with all the relationships on the same image independently, which ignore the strong logical constraints among the social relationships on an image.

\textbf{Graph Neural Networks.}
The notation of GNNs was firstly introduced by Gori \emph{et al.} \cite{gori2005new}, and further elaborated in  \cite{scarselli2008graph}.  Scarselli \emph{et al.} trained the GNNs via the Almeida-Pineda algorithm.
Li \emph{et al.} \cite{li2015gated} modified the previous GNNs to use gated recurrent units and modern optimization techniques. Inspired by the huge success of convolutional networks in the computer vision domain, many efforts have been devoted to re-defining the notation of convolution for graph data \cite{bruna2013spectral,defferrard2016convolutional,kipf2017semi}. One of the most famous works is GCN proposed by Kipf \emph{et al.} \cite{kipf2017semi}. Li \emph{et al.} \cite{li2018deeper} developed deeper insights into the GCN model and argued that if a GCN was deep with many convolutional layers, the output features might be over-smoothed. Gilmer \emph{et al.} \cite{gilmer2017neural} reformulated the existing GNN models into a single common framework  called message passing neural networks (MPNNs). Veli{\v{c}}kovi{\'c} \emph{et al.} \cite{velickovic2018graph} presented graph attention networks leveraging masked self-attentional layers. Xu \emph{et al.} \cite{xu2018powerful} characterized the discriminative power of popular GNN variants and developed a simple architecture that was provably the most expressive among the class of GNNs. Hamilton \emph{et al.} \cite{hamilton2017inductive} presented GraphSAGE  to efficiently generate node embeddings for previously unseen data. GNNs have been proven to be an effective tool for relational reasoning \cite{schlichtkrull2018modeling,Sun2019actionforecast,kipf2018neural,sun2019stochastic,tacchetti2018relational}. Schlichtkrull \emph{et al.} \cite{schlichtkrull2018modeling} proposed R-GCNs and applied them to knowledge base completion tasks. Chen \emph{et al.} \cite{Chen2019graphreason} proposed a GloRe unit for reasoning globally.

\section{Approach}
In this section, we first illustrate the importance of the paradigm shift. Then we present the details of the graph building and the proposed GR$^2$N. Finally, we provide comparisons between the proposed approach and other GNNs' methods.

\subsection{Revisiting the Paradigm of Social Relation Recognition}
Formally, the problem of social  relation recognition can be formulated as a probability function:
given an input image $\bm{I}$, bounding box values $\bm{b_i}$ and queries $\bm{x}$:
\begin{equation}
\bm{x} = \{x_{i, j} | i = 1,2,...,N,j = 1,2,...,N\}
\label{equ:query}
\end{equation}
where $x_{i, j}$ is the social relation between the person $i$ and $j$, and $N$ is the total number
of people in the image $\bm{I}$. The goal of social relation recognition is to find an optimal value of $\bm{x}$:
\begin{equation}
\bm{x^*} = \mathop{\arg\max}_{\bm{x}} P(\bm{x}|\bm{I},\bm{b}_1,\bm{b}_2,...,\bm{b}_N)
\label{equ:rightOptimal}
\end{equation}

As far as we know, all existing methods of social relation recognition organize data as person pair-based, and each
sample of these methods consists of an input image $\bm{I}$ and a target pair of people highlighted by bounding boxes
{$\bm{b_i},\bm{b_j}$}. Therefore, the actual optimization objective of these methods is as follows:
\begin{equation}
x_{i , j}^* = \mathop{\arg\max}_{x_{i, j}} P(x_{i, j}|\bm{I},\bm{b}_i,\bm{b}_j)
\label{equ:wrongOptimal}
\end{equation}

The optimization objective in  (\ref{equ:wrongOptimal}) is consistent with that in  (\ref{equ:rightOptimal})
if and only if the following equation holds:
\begin{equation}
P(\bm{x}|\bm{I},\bm{b}_1,\bm{b}_2,...,\bm{b}_N) = \prod_{1 \leq i,j \leq N} P(x_{i , j}|\bm{I},\bm{b}_i,\bm{b}_j)
\label{equ:holdcond}
\end{equation}

It is known that \eqref{equ:holdcond} holds if and only if all relations on the input image $\bm{I}$  are independent of each other.
Unfortunately, as we discussed earlier, this is not true and the relations on an image are usually highly related.
Let $\bm{y}$ be the ground truths corresponding to queries $\bm{x}$: $\bm{y} = \{y_{i, j} | i = 1,2,...,N,j = 1,2,...,N\}$.
Then the paradigm of existing methods treats quadruplets $<\bm{I},\bm{b}_i,\bm{b}_j,y_{i , j}>$ as independent samples to optimize
the objective in  (\ref{equ:wrongOptimal}), which can not lead to the optimal value of  that in  (\ref{equ:rightOptimal}).
On the contrary, we reorganize the data as image-based to directly optimize the objective formulated in  (\ref{equ:rightOptimal}) with a thorough understanding of the social scene. Our method formulates each sample as an (N+2)-tuple $<\bm{I},\bm{b}_1,\bm{b}_2,...,\bm{b}_N,\bm{y}>$ and jointly infers the relations for each image.

\subsection{From Image to Graph}

We model each person on an image as a node in a graph, and the edge between two nodes represents the relation between the
corresponding two people. Then graph operations are applied to perform joint reasoning. So the first step is to create
nodes in the graph using the input image $\bm{I}$ and bounding boxes $\bm{b}_1,\bm{b}_2,...,\bm{b}_N$. As commonly used in detection \cite{girshick2015fast,ren2015faster,redmon2016you,redmon2017yolo9000}, a convolutional neural network takes the input image $\bm{I}$ as input,
and  the features of object proposals are extracted directly from the last convolutional feature map $\bm{H}$.
More specifically, we first map the bounding boxes $\bm{b}_1,\bm{b}_2,...,\bm{b}_N$ to $\bm{b}_1^{'},\bm{b}_2^{'},...,\bm{b}_N^{'}$
according to the
transformation from $\bm{I}$ to $\bm{H}$. Then the feature $\bm{p_i}$ of person $i$ is obtained by employing an RoI pooling
layer on the feature map $\bm{H}$: $\bm{p}_i = f_{RoIPooling}(\bm{H},\bm{b}_i^{'})$. In this way, we obtain all the features of people
on the image $\bm{I}$: $\bm{p} = \{\bm{p}_1,\bm{p}_2,...,\bm{p}_N\}$, $\bm{p}_i \in \mathbb{R}^F$, where $F$ is the feature dimension for each person. Subsequently, these features are set as the initial feature representations of nodes in the graph:
$\bm{h}^0 = \{\bm{h}_1^0,\bm{h}_2^0,...,\bm{h}_N^0\}$, where $\bm{h}_i^0 = \bm{p}_i$.

\subsection{Graph Relational Reasoning Network}

Having created the nodes in the graph, we propose a graph relational reasoning network to perform relational reasoning on the graph.
We begin with a brief description of the overall framework of our GR$^2$N. Let $\mathcal{G} = (\mathcal{V},\mathcal{E})$ denotes a graph with
node feature vectors $\bm{h}_v$ for $v \in \mathcal{V}$ and edge features $\bm{e}_{vw}$ for $(v,w) \in \mathcal{E}$.
The framework of GR$^2$N  can be formulated as a message passing phase and a readout phase following  \cite{gilmer2017neural}.
The message passing phase, which is defined in terms of message function
$M_t$ and vertex update function $U_t$, runs for $T$ time steps. During the message passing phase, the aggregated messages $\bm{m}_v^{t+1}$ are used to update the hidden states $\bm{h}_v^t$:
\begin{equation}
\bm{m}_v^{t+1} = \sum_{w \in N(v)} M_t(\bm{h}_v^t, \bm{h}_w^t,\bm{e}_{vw})
\label{equ:message}
\end{equation}
\begin{equation}
\bm{h}_v^{t+1} = U_t(\bm{h}_v^t,\bm{m}_v^{t+1})
\label{equ:update}
\end{equation}
where $N(v)$ denotes the neighbors of $v$ in graph $\mathcal{G}$.
The readout phase computes feature vectors for edges using the readout function $R$.
The message function $M_t$, vertex update function $U_t$, and readout function $R$ are all differentiable functions.

In each time step of the message passing phase, the state of the node is updated according to the messages sent by its neighbor nodes, which requires us to know the topology of the entire graph. In social relation recognition, we use edges to represent social relations, and the edge type represents the category of the relation. However, the task of social relation recognition is to predict the existence of edges and the type of edges, which means that we do not know the topology of the graph at the beginning. To address this issue, the proposed GR$^2$N first models the edges in the graph.

The social relations on an image have strong \emph{logical constraints} and our GR$^2$N aims to grasp these constraints to generate a reasonable and consistent social relation graph.  As each kind of social relations has its specific logical constraints, we let GR$^2$N propagate different types of messages for different types of edges.
Specifically, we construct a virtual relation graph for each kind of relationship, in which the edge represents the existence of this kind of relationship between two nodes. To achieve mutual reasoning and information fusion among different relationships, the feature representations of nodes are shared across  all created virtual relation graphs.

Mathematically, we use K to denote the number of social relationship categories, and K virtual relation graphs $(\mathcal{V},\mathcal{E}^1),(\mathcal{V},\mathcal{E}^2),...,(\mathcal{V},\mathcal{E}^K)$ are created to model the K social relationships separately. Assuming that we have obtained the node feature representations at time step $t$: $\bm{h}^t = \{\bm{h}_1^t,\bm{h}_2^t,...,\bm{h}_N^t\}$, we first model the edge embedding in the virtual relation graph:
\begin{equation}
\bm{e}_{i,j}^{k,t+1} = f_{r_k}(\bm{h}_i^t,\bm{h}_j^t) = [\bm{W}_{r_k} \bm{h}_i^t || \bm{W}_{r_k} \bm{h}_j^t ]
\label{equ:edgeEmbeding}
\end{equation}
where $\bm{e}_{i,j}^{k,t+1} $ denotes the embedding of edge $(i,j)$ in virtual relation graph $(\mathcal{V},\mathcal{E}^k)$ at time $t+1$ and $||$ is the concatenation operation. $f_{r_k}(\cdot)$ is parameterized by a weight matrix $\bm{W}_{r_k} \in \mathbb{R}^{F \times F}$. In standard GNNs, the existence of edges is binary and deterministic. In this paper, we consider rather a probabilistic soft edge, \emph{i.e.} edge $(i, j)$ in the virtual relation graph $(\mathcal{V},\mathcal{E}^k)$ exists according to the probability:
\begin{equation}
\alpha_{i,j}^{k,t+1} = \sigma (\bm{a}_{r_k}^{\top} \bm{e}_{i,j}^{k,t+1} )
\label{equ:edgeProb}
\end{equation}
where $\bm{a}_{r_k} \in \mathbb{R}^{2F}$ is a weight vector and $\cdot ^{\top}$ denotes transposition. The sigmoid function $\sigma(\cdot)$  is employed as the activation function to normalize the values between 0 and 1. In the end, the $\mathcal{E}^k$ can be represented as the set $\{\alpha_{i,j}^{k}|1 \leq i,j \leq N\}$ if we ignore the time step.

Once obtained, we propagate messages on each virtual relation graph according to the soft edges and aggregate  messages across all virtual relation graphs:
\begin{equation}
\bm{m}_i^{t+1} = \sum_{j \in N(i)} \sum_{k =1}^K \alpha_{i,j}^{k,t+1} f_m(\bm{h}_i^t, \bm{h}_j^t,\bm{e}_{i,j}^{k,t+1})
\label{equ:message1}
\end{equation}
For simplicity, we reuse the previous weight $\bm{W}_{r_k}$ and let $f_m(\bm{h}_i^t, \bm{h}_j^t,\bm{e}_{i,j}^{k,t+1}) = \bm{W}_{r_k} \bm{h}_j^t$. So we reformulate $\bm{m}_i^{t+1}$ as:
\begin{equation}
\bm{m}_i^{t+1} = \sum_{j \in N(i)} M_t(\bm{h}_i^t, \bm{h}_j^t,\bm{e}_{i,j}^{k,t+1}) = \sum_{j \in N(i)} (\sum_{k =1}^K \alpha_{i,j}^{k,t+1} \bm{W}_{r_k} \bm{h}_j^t)
\label{equ:message2}
\end{equation}

Finally, we use the aggregated messages to update the hidden state of node:
\begin{equation}
\bm{h}_i^{t+1} = U_t(\bm{h}_i^t,\bm{m}_i^{t+1}) = f_{GRU}(\bm{h}_i^t,\bm{m}_i^{t+1})
\label{equ:nodeUpdate}
\end{equation}
where $f_{GRU}(\cdot)$ is the Gated Recurrent Unit update function introduced in \cite{cho2014learning}.

Having repeated the above process for $T$ time steps, we obtain the final node feature representations:
$\bm{h}^T = \{\bm{h}_1^T,\bm{h}_2^T,...,\bm{h}_N^T\}$. Then the readout function $R$ is applied to the features $\bm{h}^T$ for  social relation recognition. As mentioned above, the task of social relation recognition is to predict the existence and type of edges in the graph, which is exactly what virtual social graphs accomplish. So the readout function $R$ simply reuses the functions that create the virtual social graphs:
\begin{equation}
\hat{x}_{i,j}^{k} = R(\bm{h}_i^T,\bm{h}_j^T,k) = \sigma(\bm{a}_{r_k}^{\top} [\bm{W}_{r_k} \bm{h}_i^T || \bm{W}_{r_k} \bm{h}_j^T ])
\label{equ:readout}
\end{equation}
where $\hat{x}_{i,j}^{k}$  indicates the probability of the existence of the $k$-th social relation between person $i$ and  $j$.
During the test stage, the social relation between person $i$ and $j$ is chosen as the category with the greatest probability: $ \mathop{\arg \max}_{k} \hat{x}_{i,j}^{k}$.

The overall framework including the backbone CNN and the proposed GR$^2$N is jointly optimized end-to-end. If we extend the ground truths $\bm{y}$ to one hot, then the loss function for the sample $<\bm{I},\bm{b}_1,\bm{b}_2,...,\bm{b}_N,\bm{y}>$ is:
\begin{equation}
\mathcal{L} = \sum_{1 \leq i,j \leq N} \sum_{k=1}^{K} -[y_{i,j}^{k} \log(\hat{x}_{i,j}^{k}) + (1 - y_{i,j}^{k}) \log(1-\hat{x}_{i,j}^{k})]
\label{equ:loss}
\end{equation}
where $y_{i,j}^{k}$ is the $k$-th element of one hot form of $y_{i,j}$.

\subsection{Discussion}
Another intuitive way to jointly model all the social relationships among people on an image is to use some common variants of GNNs, such as GGNN \cite{li2015gated} and GCN \cite{kipf2017semi}. They can be utilized to perform graph operations on the graphs to replace the proposed GR$^2$N. In addition, GNNs are used in some other tasks such as group activity recognition \cite{deng2016structure}, and scene graph generation \cite{xu2017scene} to model the relationships between objects. In the following, we will elaborate on the key difference between the proposed GR$^2$N and those GNNs' methods.

Different from group activity recognition, scene graph recognition, and other tasks, different kinds of social relations have \emph{strong logical constraints} in the social relation recognition task. For example, knowing that A-C is mother-child and B-C is father-child, we can directly infer that A-B is spouses even without checking the relation A-B itself. However, most existing methods of jointly reasoning all relations such as GGNN \cite{li2015gated}, GCN \cite{kipf2017semi}, Structure Inference Machines \cite{deng2016structure} and Iterative Message Passing \cite{xu2017scene}  simply exploit \emph{contextual information} via message passing, which fails to \emph{explicitly} grasp the \emph{logical constraints} among different types of social relations. Instead, our GR$^2$N constructs a virtual social relation graph for each social relation type, where each graph learns a \emph{type-specific} message passing matrix $\bm{W}_{r_k}$. As each type of social relation has its \emph{specific logical constraints}, the proposed GR$^2$N is aware of \emph{relation types} and better exploits the \emph{type-specific logical constraints} in social relation recognition. In this way, our method can better perform relational reasoning and generate a reasonable and consistent social relation graph.

\section{Experiments}
In this section, we first introduce the datasets and present some implementation details of our approach. Then we evaluate the performance of our method using  quantitative and qualitative analysis.

\subsection{Datasets}
We evaluated the GR$^2$N and existing competing methods on the People in Photo Album (PIPA) \cite{sun2017domain} database and People in Social Context (PISC) database \cite{li2017dual}.

\textbf{PIPA:} The PIPA database is collected from Flickr photo albums for the task of person recognition \cite{zhang2015beyond}. Then the dataset is extended by Sun \emph{et al.} with 26,915 person pair annotations based on the social domain theory \cite{bugental2000acquisition}.
The PIPA dataset involves two-level recognition tasks: 1) social domain recognition focuses on five categories of social domain, i.e., attachment domain, reciprocity domain, mating domain, hierarchical power domain, and coalitional group domain; 2) social relationship recognition focuses on 16 finer categories of relationship, such as friends, classmates, father-child, leader-subordinate, band members, and so on. For fair comparisons, we adopt the standard train/val/test split released by Sun \emph{et al.} \cite{sun2017domain}, which uses 13,672 domain/relation instances in 5,857 images for training, 709 domain/relation instances in 261 images for validation,  and  5,106 domain/relation instances in 2,452 images for testing. The top-1 classification accuracy is used for evaluation. 

\textbf{PISC:} The PISC database collects 22,670 images and is annotated following the relational model theory \cite{fiske1992four}. It has hierarchical social relationship categories: coarse-grained relationships (intimate, not-intimate and no relation) and fine-grained relationships (commercial, couple, family, friends, professional and no-relation). For coarse-grained relationship, 13,142 images with 49,017 relationship instances are used for training, 4,000 images with 14,536 instances are used for validation, 4,000 images with 15,497 instances are used for testing. For fine-grained relationship, the train/val/test set has 16,828 images and 55,400 instances, 500 images and 1,505 instances, 1250 images and 3,961 instances, respectively. We follow the commonly used metrics \cite{Arushi2019social,li2017dual,Wang2018Deep} on the PISC dataset and report the per-class recall for each relationship and the mean average precision (mAP) over all relationships.

\subsection{Implementation Details}
For fair comparisons, we employed ResNet-101 as the backbone CNN following \cite{li2017dual,Wang2018Deep}. The ResNet-101 was initialized with the pre-trained weights on ImageNet \cite{russakovsky2015imagenet}. The shape of the output region of the RoI pooling layer was set to be the same as the shape of the last convolution feature map $\bm{H}$, and then a global average pooling layer was used to obtain a 2048-D feature for each person. The time step $T$ was set to 1, which achieved the best result. The model was trained by Adam optimization \cite{kingma2014adam} with a learning rate of $10^{-5}$. We trained our method for 10 epochs with a batch size of 32. During training, images were horizontally flipped with probability 0.5 and randomly cropped for data augmentation.

In a mini-batch, different images might have different numbers of people, which made the size of the graphs variable. To deal with this problem, we set the number of people $N$ on an image  to the maximum possible persons $N_{max}$.
If the number of people on an image was less than $N_{max}$, then we set the missing nodes as empty ones and there would be no soft edges between the empty nodes and the real nodes. For those relationships that were not labeled, no loss would occur from them in (\ref{equ:loss}) to avoid having an impact on network training. Another issue was that the social relationships on the PISC dataset were highly imbalanced, especially for the fine-level relationships. To address this, we adopted the reweighting strategy for fine-grained relationship recognition on the PISC dataset. Specifically, the samples were reweighted inversely proportionally to the class frequencies.

\subsection{Results and Analysis}

\textbf{Comparisons with the State-of-the-Art Methods.}
We compare the proposed approach with several existing state-of-the-art methods. The details of these competing methods are as follows:

\emph{Finetuned CNN + SVM} \cite{sun2017domain}. Double-stream CaffeNet is used to extract features, then the extracted features are utilized to train a linear SVM.

\emph{All attributes + SVM} \cite{sun2017domain}. Many semantic attribute categories including age, gender, location, and activity are used in this method. Then all attribute features are concatenated to train a linear SVM.

\emph{Pair CNN} \cite{li2017dual}. This model consists of two  CNNs (ResNet-101) with  shared weights. The input is two cropped image patches for the two individuals and the extracted features are concatenated for social relation recognition.

\emph{Dual-Glance} \cite{li2017dual}. Two cropped individual patches and the union region of them are sent to CNNs to extract features. The introduced second glance exploits surrounding proposals as contextual information to refine the predictions.

\emph{SRG-GN} \cite{Arushi2019social}. Five CNNs are utilized to extract scene and attribute context information (\emph{e.g.}, age, gender, and activity). Then these features are utilized to update the state of memory cells like GRUs.

\emph{MGR} \cite{zhang2019multi}. It employ five CNNs to extract the global, middle, and fine granularity features to comprehensively capture the multi-granularity semantics.

\emph{GRM} \cite{Wang2018Deep}. It replaces the second glance in Dual-Glance with a pre-trained Faster-RCNN detector \cite{ren2015faster} and  a Gated Graph Neural Network \cite{li2015gated} to model the interaction between the contextual objects and the persons of interest.

\begin{table}[t]
  \caption{Comparisons of the accuracy between our GR$^2$N and other state-of-the-art methods on the PIPA dataset.}
  \label{table:stateofthearts}
  \renewcommand\tabcolsep{4pt}
  \centering
  \begin{tabular}{lcccc}
    \toprule
    \multirow{2}*{methods} & \# of input  & \multirow{2}*{\# of CNNs} & \multirow{2}*{domain} & \multirow{2}*{relation} \\
    ~ & image patches & ~ & ~ & ~ \\
    \midrule
    Finetuned CNN + SVM \cite{sun2017domain}& 2 & 2 & 63.2\% & 48.6\% \\
    All attributes + SVM \cite{sun2017domain} & 4 & 10 & 67.8\% & 57.2\% \\
     Pair CNN \cite{li2017dual}& 2 & \textbf{1} & 65.9\% & 58.0\% \\
     Dual-Glance  \cite{li2017dual}  & 4 & 3 & - & 59.6\% \\
     SRG-GN \cite{Arushi2019social} & 3 & 5 & - & 53.6\% \\
     GRM \cite{Wang2018Deep} & 4 & 3 &   - & 62.3\% \\
     MGR \cite{zhang2019multi} & 4 & 5 & - & \textbf{64.4\%} \\
     \midrule
     GR$^2$N-224 & \textbf{1} & \textbf{1} & 69.3\% & 61.3\% \\
     GR$^2$N-448 & \textbf{1} & \textbf{1}& \textbf{72.3\%} & \textbf{64.3\%} \\
    \bottomrule
  \end{tabular}
\end{table}

All of the above methods are person pair-based, which means that they consider the social relations on the same image separately. It is worth noting that although the SRG-GN mentions that they can generate a social relationship graph,  they actually still process each relationship independently during the training and testing phases. These methods usually crop the image patches for interested individuals and resize them to 224 $\times$ 224 pixels while our approach takes the entire image as input. In order to make a fair comparison, we should choose the appropriate entire input image size to ensure that the personal area of interest is roughly equal to 224 $\times$ 224. In the PIPA and PISC dataset, the area of the bounding boxes is on average 1/4 and 1/5 of the area of the images, respectively, so we resize the original image to 448 $\times$ 448 pixels, which is denoted as GR$^2$N-448.
Although this is still unfavorable to our approach on the PISC datasets, our method can still achieve promising performance.
Besides, we also report the performance of GR$^2$N-224 to show the effectiveness of our method, which resizes the original image to 224 $\times$ 224 pixels.

\begin{table}[t]
  \caption{We present the per-class recall
for each relationship and the mAP over all relationships (in \%) on the PISC dataset. The first and second best scores are highlighted in red and blue colors, respectively. (Int: Intimate, Non: Non-Intimate, NoR: No Relation, Fri: Friends, Fam: Family, Cou: Couple, Pro: Professional, Com: Commercial)}
  \label{table:PISCSoTA}
  \renewcommand\tabcolsep{3pt}
  \centering
  \begin{tabular}{l|ccc|c|cccccc|c}
    \toprule
    \multirow{2}*{Methods}   & \multicolumn{4}{c|}{Coarse relationships} & \multicolumn{7}{c}{Fine relationships}\\
    \cmidrule{2-12}
    ~  & Int & Non & NoR & mAP & Fri & Fam & Cou & Pro & Com & NoR  & mAP \\
    \midrule
     Pair CNN \cite{li2017dual} & 70.3 & 80.5 & 38.8 &  65.1 &  30.2 &  59.1 & 69.4 & 57.5 & 41.9 & 34.2 &  48.2 \\
     Dual-Glance \cite{li2017dual}  &  73.1 &  84.2 &  59.6 &  79.7 &  35.4&  68.1&  76.3&  70.3 & 57.6 & 60.9&  63.2 \\
     GRM \cite{Wang2018Deep} & 81.7  &73.4 & 65.5  & 82.8 & 59.6 & 64.4  &58.6 & 76.6 & 39.5  &67.7  &68.7 \\
     MGR \cite{zhang2019multi} & - & - & - & - & 64.6 & 67.8 & 60.5 &  76.8 & 34.7 & 70.4 & 70.0 \\
     SRG-GN \cite{Arushi2019social}  & - & - & - & - & 25.2 & 80.0 & 100.0 & 78.4  & 83.3 & 62.5 & 71.6 \\
     \midrule
     GR$^2$N-224  & 76.3 & 65.3 & 74.0 & 72.2 &  54.7 & 68.5 & 78.1 & 78.0 & 49.7 & 57.5 & 68.6  \\
     GR$^2$N-448  & 81.6 & 74.3 & 70.8 & \textbf{83.1} & 60.8 & 65.9 & 84.8 & 73.0 & 51.7 & 70.4 & \textbf{72.7}  \\
    \bottomrule
  \end{tabular}
\end{table}

The experimental results of  social domain recognition and social relationship recognition  on the PIPA database are shown in Table \ref{table:stateofthearts}. We first compare our method with a simple baseline Pair CNN, which, like our method, does not use any scene and attribute context cues, so we can see the benefits brought by the paradigm shift. We observe that our method significantly outperforms Pair CNN. Specifically, the GR$^2$N-448 achieves an accuracy of 64.3\% for social relation recognition and 72.3\% for social domain recognition, improving the Pair CNN by 6.3\% and 6.4\% respectively.  What's more, even the GR$^2$N-224 with inferior input image size outperforms the Pair CNN by 3.3\% and 3.4\% for  social relation recognition and social domain recognition respectively, which demonstrates that the paradigm shift and GR$^2$N's superior relational reasoning ability bring significant performance gains.

Next, we compare our method with state-of-the-art methods. In terms of social relation recognition, the proposed GR$^2$N-448 improves GRM by 2.0\%. What is worth mentioning is that GRM uses three individual CNNs to extract features from four image patches including one 448 $\times$ 448 entire image and three 224 $\times$ 224 person patches to exploit key contextual cues, while  the GR$^2$N only uses one input  image and one convolutional neural network. Our method also achieves competitive results with MGR whereas MGR uses 5 CNNs and 4 patches. For social domain recognition, our method GR$^2$N-448 achieves an accuracy of 72.3\%, which improves the performance of All attributes + SVM by 4.5\% without using any semantic attribute categories.
Clearly, these results illustrate the superiority of our method.

Table \ref{table:PISCSoTA} shows the experimental comparison with the recent state-of-the-art methods on the PISC database. We observe that our method achieves an mAP of 83.1\% for the coarse-level recognition and 72.7\% for the fine-level recognition, improving the simple baseline Pair CNN by a large margin: 18.0\% for coarse relationships and 24.5\% for fine relationships, which further validates the advantage of the paradigm shift.  Our approach outperforms SRG-GN by 1.1\% with a much simpler framework (1 neural net vs. 5 neural nets, 1 image patch vs. 3 image patches). Compared with GRM, GR$^2$N-448 achieves competitive performance for coarse relationship recognition and superior performance for fine relationship recognition with only one image patch and one neural net, while the GRM uses four image patches and three neural nets. It is worth noting that the experimental results of our approach on the PISC dataset  are achieved with inferior input image size, which further illustrates the effectiveness of our GR$^2$N.

\textbf{Runtime Analysis.} In addition to being \emph{simpler} and \emph{more accurate}, another advantage of our GR$^2$N is that it is \emph{faster}. Since our method handles all relationships on an image at the same time, the GR$^2$N is computationally efficient.
We conduct experiments under different batch sizes to compare the speed of different methods using a GeForce GTX 1080Ti GPU on the PIPA dataset. To be fair, the version of GR$^2$N used here is GR$^2$N-448.
The results of the forward runtime (seconds/image) are reported in Table \ref{table:speed}. The * in the table represents memory overflow. Compared with Pair CNN, our method achieves about  $2 \times$ speed-ups, which shows the benefits of the paradigm shift. The GR$^2$N-448 is $4 \times  \sim  7 \times$ faster than the state-of-the-art method GRM, which further demonstrates the \emph{efficiency} of our approach.
Since the average number of bounding boxes per image on the PIPA datasets is only 2.5, we expect higher speedup when there are more individuals on an image.

\begin{table}[t]
\parbox{.65\linewidth}{
\centering
\renewcommand\tabcolsep{5pt}
\caption{Comparisons of the runtime (seconds/image) under different batch sizes for social relation recognition on the PIPA dataset.}\label{table:speed}
  \begin{tabular}{lcccc}
    \toprule
    \multirow{2}*{methods} & \multicolumn{4}{c}{batch size} \\
    \cmidrule{2-5}
    ~ & 1 & 2 & 4 & 8 \\
    \midrule
     GRM & 0.294  & 0.171 &   0.089 & *  \\
     Pair CNN & 0.077 & 0.045 & 0.039 & 0.037  \\
     GR$^2$N-448 & \textbf{0.046} & \textbf{0.025}& \textbf{0.021} & \textbf{0.021}   \\
    \bottomrule
  \end{tabular}
}
\hfill
\parbox{.33\linewidth}{
\centering
\renewcommand\tabcolsep{5pt}
\caption{Comparisons with GCN and GGNN for social relation
recognition.}\label{table:ablation}
 \begin{tabular}{lc}
    \toprule
    methods & accuracy \\
    \midrule
    Pair CNN & 58.0\% \\
    \midrule
     GCN   & 59.3\%   \\
     GGNN   & 59.8\%  \\
     GR$^2$N & \textbf{61.3\%} \\
    \bottomrule
  \end{tabular}
}
\end{table}

\textbf{Ablation Study.}
 To validate the effectiveness of the proposed GR$^2$N, we compare it with two commonly used GNN variants: GGNN \cite{li2015gated} and GCN \cite{kipf2017semi}. To apply GGNN and GCN, we connect the nodes together into a fully connected graph. Following \cite{wang2018videos}, we also add a residual connection in every layer of GCN, which observes an improvement in  performance in the experiments. For a fair comparison, all three methods use an image of size 224 $\times$ 224 as input. The readout function of GGNN and GCN is a two-layer neural network and the edge embedding is obtained by concatenating the features of the corresponding two nodes. The cross-entropy loss function is used to train GGNN and GCN.
\begin{figure}[t]
  \centering
  \includegraphics[width=1.0\linewidth]{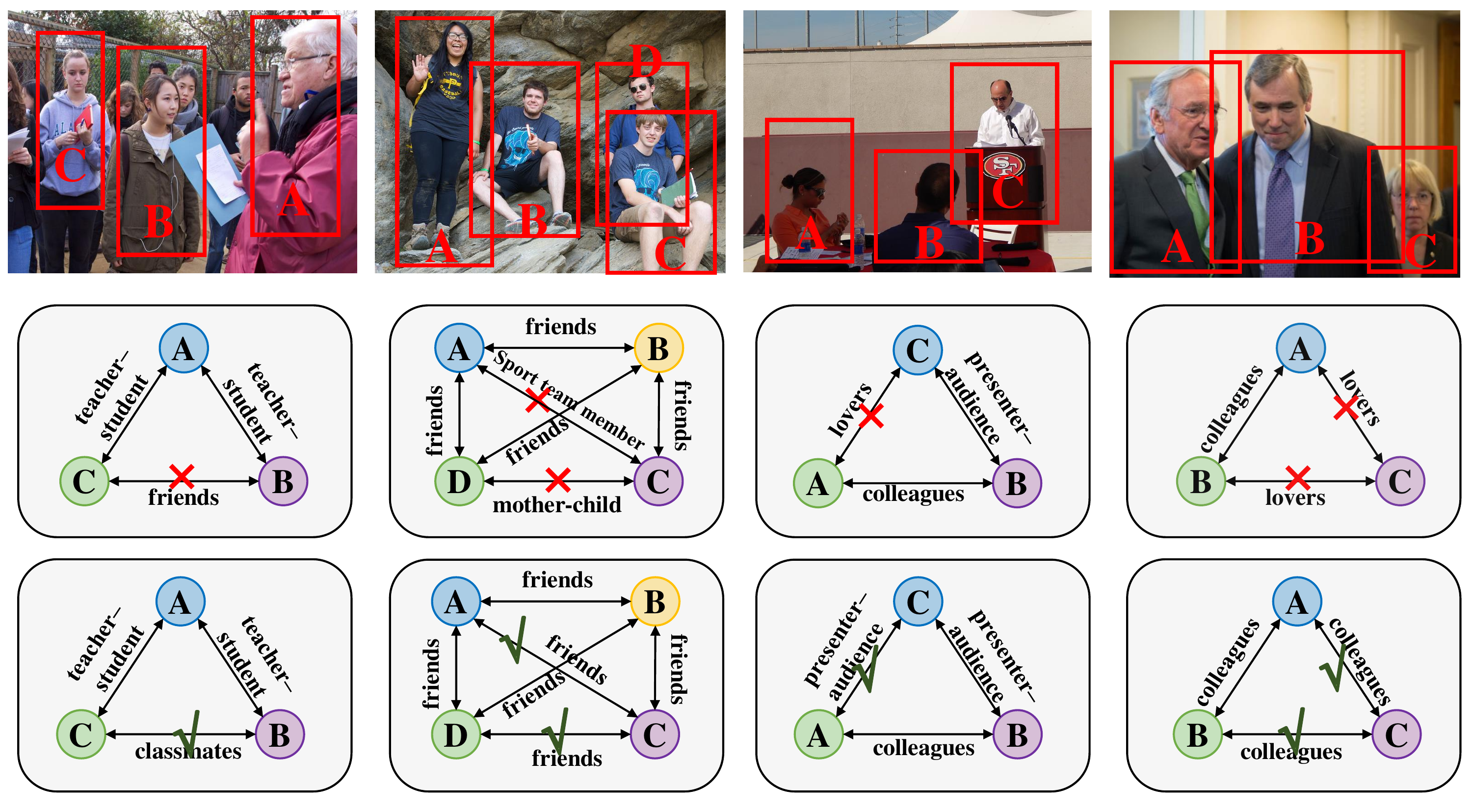}
  \caption{Some examples of qualitative evaluation. The first row shows the input images, the second row is the corresponding predicted results of Pair CNN, and the third row is the corresponding predicted results of our method. The red cross mark indicates a prediction error, while the green check mark indicates that the corresponding  error is corrected. We see that our method corrects the prediction errors by performing graph-based relational reasoning.}
  \label{fig:Qualitative}
\end{figure}

The comparisons on the PIPA dataset by accuracy are listed in Table \ref{table:ablation}. We observe that our proposed GR$^2$N outperforms GCN and GGNN, which demonstrates that the proposed GR$^2$N effectively performs relational reasoning and exploits the strong logical constraints of social relations on the same image. On the other hand, all three methods adopt the image-based paradigm and we see that they all outperform Pair CNN. In fact, these three methods achieve superior or competitive performance compared with some of the state-of-the-art methods which adopt the person pair-based paradigm, such as Dual-Glance and SRG-GN. Note that these GNNs' methods only use one input patch and one convolutional neural net, which illustrates the superiority of the image-based paradigm over the person pair-based paradigm.

\textbf{Qualitative Evaluation.}
\begin{figure}[t]
  \centering
  \includegraphics[width=0.8\linewidth]{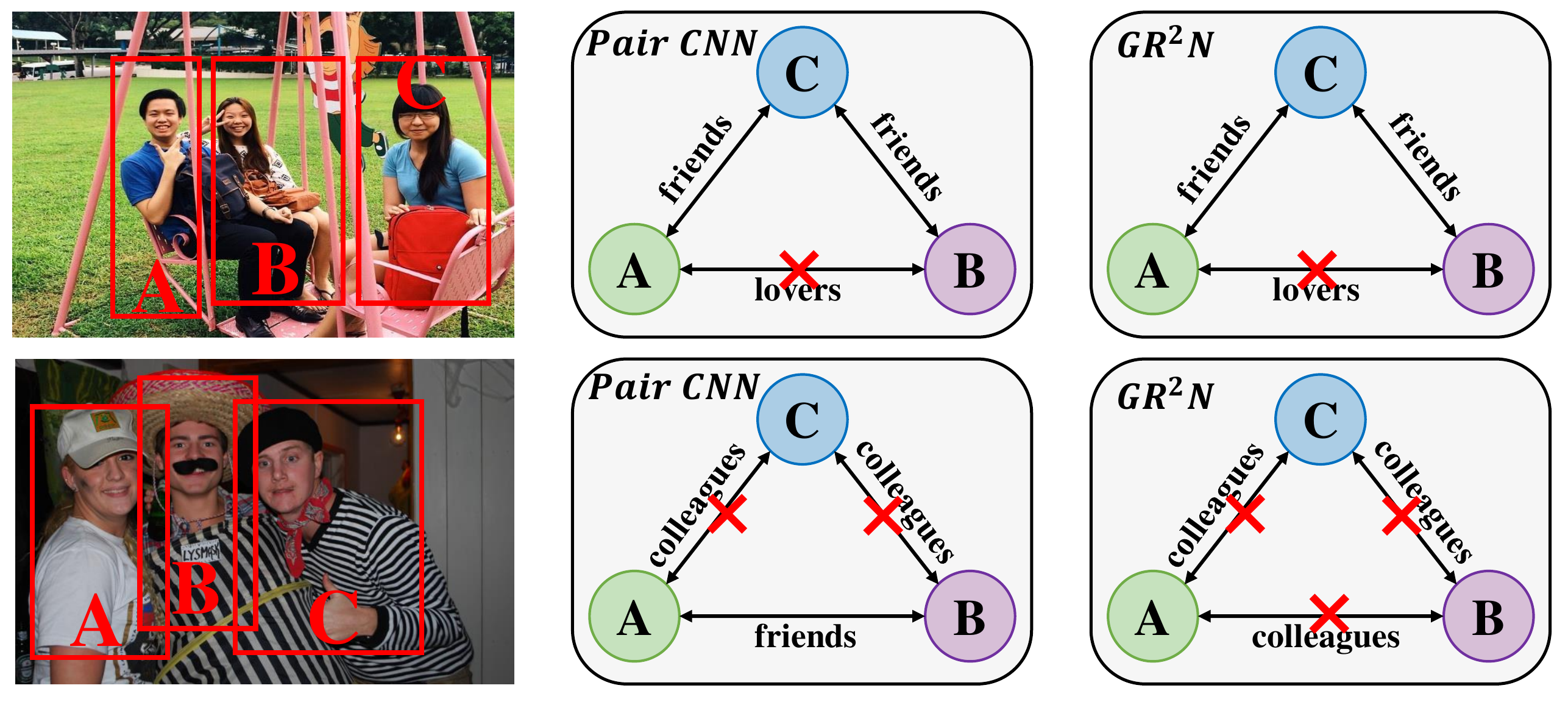}
  \caption{Some failure cases of our method on the PIPA dataset. The first column shows the input images, the second column is the corresponding predicted results of Pair CNN, and the third column is the corresponding predicted results of our method.  The red cross mark indicates a prediction error.}
  \label{fig:Fcase}
\end{figure}
In this part, we present some examples to illustrate how our GR$^2$N improves  performance by performing relational reasoning and exploiting the strong logical constraints of social relations on a graph in Fig. \ref{fig:Qualitative}. For the first example, Pair CNN independently processes the relations on the image and predicts that the relation between B and C is friends. On the contrary, our method jointly infers all relations for each image. We easily infer that the relation between B and C is classmates when we know that the relation between A and B and  the relation between A and C both are teacher-student. This is also the case in the rest of the examples. When the relationships on an image are processed independently, some obviously unreasonable and contradictory relationships may occur simultaneously, and our method is capable of  correcting them through relational reasoning with a thorough understanding of the social scene.
Clearly, our method  better models the interactions among people on the graph, and generates a reasonable and consistent social relation graph. We also present some failure cases in Fig. \ref{fig:Fcase}. We see that our method can not correct the prediction error in the first case since the error does not cause any obvious conflict. Although the ground truth of the relationship between person A and B is friends, predicting it as lovers does not lead to a contradictory social relation graph. Our method may also fail when most predictions are wrong and our model performs relational reasoning based on these false predictions, as shown in the second case.

\section{Conclusion}

In this paper, we have presented the GR$^2$N for social relation recognition, which is a \emph{simpler}, \emph{faster}, and \emph{more accurate} method. Unlike existing methods, we simultaneously reason all relations for each image and treat images as independent samples rather than person-pairs. Our method constructs a graph for each image to model the interactions among people and perform relational reasoning on this graph with fully exploiting the  strong logical constraints  of relations. Experimental results demonstrate that our method generates a reasonable and consistent social relation graph. Moreover, GR$^2$N achieves better performance with less time cost compared with the state-of-the-art methods, which further illustrates the \emph{effectiveness} and \emph{efficiency} of our method.

\section*{Acknowledgments}
This work was supported in part by the National Key Research and Development Program of China under Grant 2017YFA0700802, in part by the National Natural Science Foundation of China under Grant 61822603, Grant U1813218, Grant U1713214, and Grant 61672306, in part by Beijing Natural Science Foundation under Grant No. L172051, in part by Beijing Academy of Artificial Intelligence (BAAI), in part by a grant from the Institute for Guo Qiang, Tsinghua University, in part by the Shenzhen Fundamental Research Fund (Subject Arrangement) under Grant JCYJ20170412170602564, and in part by Tsinghua University Initiative Scientific Research Program.
%
%
\bibliographystyle{splncs04}
\bibliography{egbib}

\newpage
\appendix
\begin{center}
\noindent{\textbf{\large{Supplementary Materials}}}
\end{center}

\section{Pipeline}
\begin{figure}[h]
  \centering
  \includegraphics[width=1.0\linewidth]{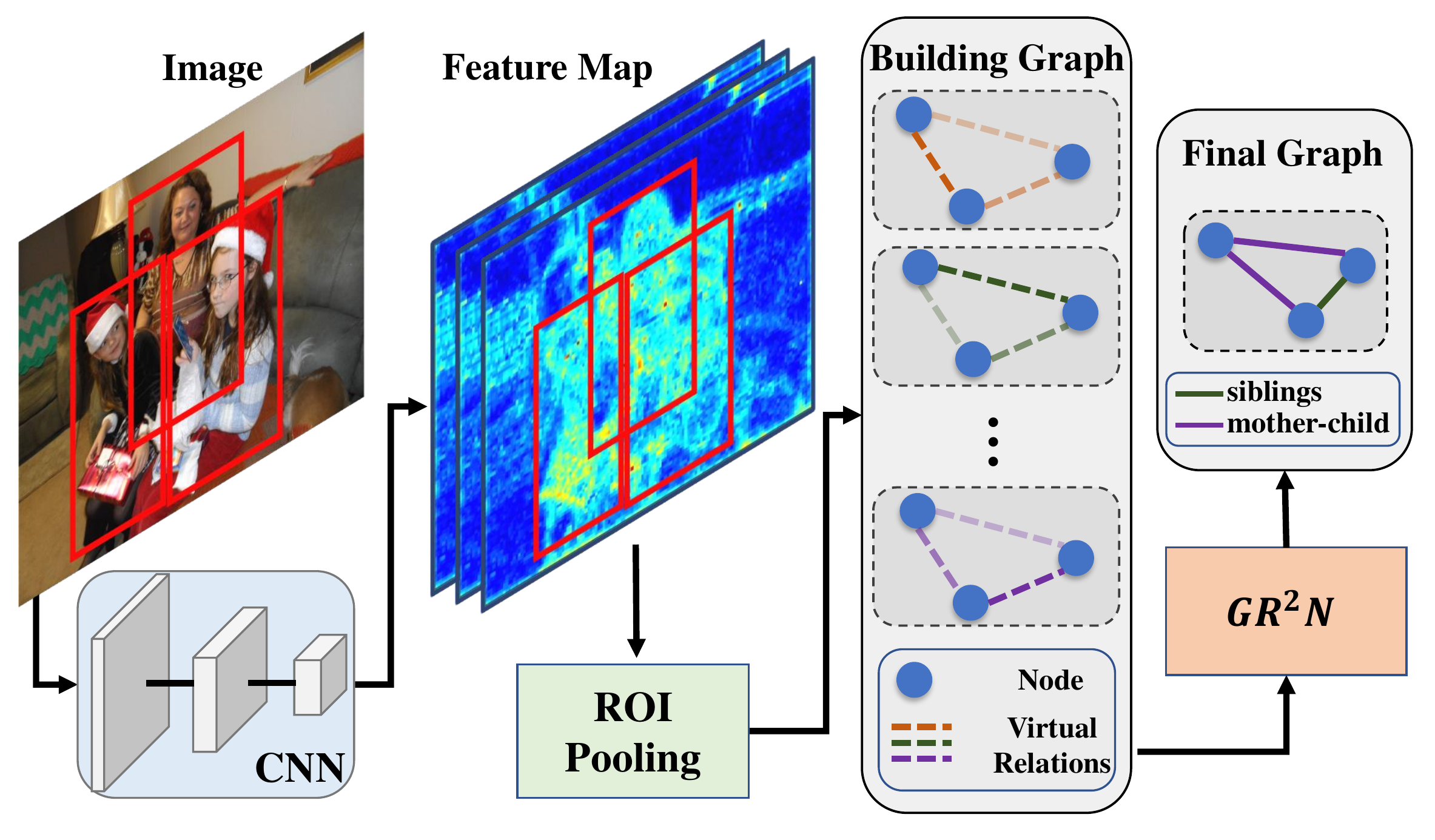}
  \caption{ An overall pipeline of our proposed method. For a given image, we first use a CNN to extract features. Then the features of people in the image are obtained from the last shared feature map using an RoI pooling layer. These features are set as the initial node embeddings. Several virtual relation graphs with shared node representations are constructed to exploit type-specific logical constraints. Finally, the proposed GR$^2$N is used to generate a reasonable  and consistent social relation graph. }
  \label{fig:framework}
\end{figure}

Fig. \ref{fig:framework} depicts the overall pipeline of our proposed GR$^2$N.

\section{Convergence}

To show the convergence of our method, we list the numerical results of GR$^2$N on the PIPA dataset in Table \ref{table:convergece}. We observe that our method eventually converges.

\begin{table}[h]
  \caption{We show how the loss function value of our method changes during the training stage as the number of iterations increases.}
  \label{table:convergece}
  \renewcommand\tabcolsep{7pt}
  \centering
  \begin{tabular}{lccccccc}
    \toprule
    Iters & 0 & 10 & 50 & 100 & 500 & 1000 & 5000 \\
    \midrule
    Loss & 0.676 & 0.520 & 0.374 & 0.258 & 0.144 & 0.102 & 0.087 \\
    \bottomrule
  \end{tabular}
\end{table}

\end{document}